\documentclass[10pt,letterpaper,conference]{ieeeconf}
\usepackage{epsfig}
\usepackage{psfrag,graphicx,color}
\usepackage{amsmath}
\usepackage{algorithm2e}
\usepackage[mathscr]{eucal}
\usepackage{epstopdf}
\usepackage{booktabs}
\usepackage{hyperref}

\IEEEoverridecommandlockouts
\usepackage[normalem]{ulem}
\usepackage{amsmath,amsfonts}
\usepackage{epsfig,psfrag,graphicx,color}

\font\bfmath=cmmib10
\mathchardef\Gamma="7100
\mathchardef\Delta="7101
\mathchardef\Theta="7102
\mathchardef\Lambda="7103
\mathchardef\Xi="7104
\mathchardef\Pi="7105
\mathchardef\Sigma="7106
\mathchardef\Upsilon="7107
\mathchardef\Phi="7108
\mathchardef\Psi="7109
\mathchardef\Omega="710A

\mathchardef\alpha="710B
\mathchardef\beta="710C
\mathchardef\gamma="710D
\mathchardef\delta="710E
\mathchardef\epsilon="710F
\mathchardef\zeta="7110
\mathchardef\eta="7111
\mathchardef\theta="7112
\mathchardef\iota="7113
\mathchardef\kappa="7114
\mathchardef\lambda="7115
\mathchardef\mu="7116
\mathchardef\nu="7117
\mathchardef\xi="7118
\mathchardef\pi="7119
\mathchardef\rho="711A
\mathchardef\sigma="711B
\mathchardef\tau="711C
\mathchardef\upsilon="711D
\mathchardef\phi="711E
\mathchardef\chi="711F
\mathchardef\psi="7120
\mathchardef\omega="7121
\mathchardef\epsilon="7122

\mathchardef\varepsilon="7122
\mathchardef\vartheta="7123
\mathchardef\varpi="7124
\mathchardef\varrho="7125
\mathchardef\varsigma="7126
\mathchardef\varphi="7127
\mathchardef\imath="717B
\mathchardef\jmath="717C

\def\bfq{{\mbox{\boldmath $q$}}}

\def\bfI{{\mbox{\boldmath $I$}}}
\def\bfJ{{\mbox{\boldmath $J$}}}
\def\bfK{{\mbox{\boldmath $K$}}}

\def\bfN{{\mbox{\boldmath $N$}}}

\def\bfsigma{{\mbox{\boldmath $\sigma$}}}

\def\smallbfW{{\raise1.5pt\hbox{\mbox{\boldmath $_W$}}}}

\def\mypsfrag#1#2#3#4#5{
        \begin{figure}[htp]
           \begin{center}
              {\leavevmode
                 {\includegraphics[width=#1truecm]{#2.eps}}
              }
           \end{center}
           \vspace{#3}
           \caption{#4}
           \vspace{-10pt}
           \label{#5}
        \end{figure}
}

\def\mydoublerowpsfrag#1#2#3#4#5#6{

\begin{figure}[htp]
        \begin{center}
        \begin{tabular}[h]{c c}
              {\leavevmode
                 {\includegraphics[width=#1truecm]{#2.eps}}
              }
              \\ 
              {\leavevmode
                 {\includegraphics[width=#3truecm]{#4.eps}}
              }
        \end{tabular}
        \vspace{-10pt}
           \caption{#5}
           \label{#6}
        \end{center}
        \end{figure}

}

\def\my4psfrag#1#2#3#4#5#6#7#8{
        \begin{figure}[htp]
        \begin{center}
	        \begin{tabular}[h]{c c}
              {\leavevmode{\includegraphics[width=#1truecm]{#2.eps}}}
              &
              {\leavevmode{\includegraphics[width=#1truecm]{#3.eps}}} \\
              {\leavevmode{\includegraphics[width=#1truecm]{#4.eps}}}
              &
              {\leavevmode{\includegraphics[width=#1truecm]{#5.eps}}}
   	     \end{tabular}
           \vspace{#6}
           \caption{#7}
           \label{#8}
        \end{center}
        \end{figure}
}

\def\mydouble4psfrag#1#2#3#4#5#6#7#8{
        \begin{figure*}[htp]
        \begin{center}
            \begin{tabular}[h]{c c}
              {\leavevmode{\includegraphics[width=#1truecm]{#2.eps}}}
              &
              {\leavevmode{\includegraphics[width=#1truecm]{#3.eps}}} \\
              {\leavevmode{\includegraphics[width=#1truecm]{#4.eps}}}
              &
              {\leavevmode{\includegraphics[width=#1truecm]{#5.eps}}}
         \end{tabular}
           \vspace{#6}
           \caption{#7}
           \label{#8}
        \end{center}
        \end{figure*}
}

\input{texdraw}

\title{\LARGE \bf Handling robot constraints within a \\ Set-Based Multi-Task Priority Inverse Kinematics Framework}
\author{Paolo Di Lillo, Stefano Chiaverini, Gianluca Antonelli
\thanks{Authors are with the Department of Electrical and Information Engineering of the University of Cassino and Southern Lazio,
		Via G. Di Biasio 43, 03043 Cassino (FR), Italy
		{\tt\small \{pa.dilillo, chiaverini, antonelli\}@unicas.it}}}
\date{August 2018}

\begin{document}

\maketitle

\begin{abstract}

Set-Based Multi-Task Priority is a recent framework to handle inverse kinematics for redundant structures. Both equality tasks, i.e., control objectives to be driven to a desired value, and set-bases tasks, i.e., control objectives to be satisfied with a set/range of values can be addressed in a rigorous manner within a priority framework. In addition, optimization tasks, driven by the gradient of a proper function, may be considered as well, usually as lower priority tasks.
In this paper the proper design of the tasks, their priority and the use of a Set-Based Multi-Task Priority framework is proposed in order to handle several constraints simultaneously in real-time. It is shown that safety related tasks such as, e.g., joint limits or kinematic singularity, may be properly handled by consider them both at an higher priority as set-based task and at a lower within a proper optimization functional. Experimental results on a 7DOF Jaco$^2$ arm with and without the proposed approach show the effectiveness of the proposed method.
\end{abstract}

\section{Introduction}

Robotic systems are requested to perform more and more complex operations in all kind of environments, leading to flexible control architectures that allow them to adapt to the particular situation in a reactive manner. The most widely used approach is to split the entire operation in several elementary control objectives, implemented as sub-tasks, possibly to be performed simultaneously. The potential conflict among tasks is resolved by setting a priority and computing the resulting motion commands that assure the achievement of the higher-priority tasks, if feasible,  and tries to accomplish the lower-priority ones as much as possible given the constraints imposed by the more important tasks. This approach has been widely applied exploiting the system redundancy and the null-space projection in both dynamic~\cite{dietrich_etal2015, dietrich2013multi} and kinematic~\cite{siciliano1990kinematic} control architectures. 

A first classification among tasks can be made with respect to the control objective that they express: {\textit{equality-based}} tasks aim to bring the task to a specific desired value, for instance to move the end-effector of a manipulator to a certain position and orientation. Most of the the main redundancy resolution algorithms in literature have been developed to handle this kind of tasks \cite{slotine1991general, nakamura1987task, Chi_tra97}. {\textit{Set-based}} tasks, or {\textit{inequality constraints}} are tasks in which the control objective is to keep the task value in an interval, i.e., above a lower threshold and below an upper threshold. In this category lie tasks such as the obstacle avoidance, the joint limit avoidance and the arm manipulability. Currently one the most popular approaches to handle this kind of tasks is to express the inverse kinematics problem as a sequence of QP (Quadratic Programming) problems \cite{Lamiraux_tro2011, mansard_ijrr14}. Task-priority frameworks have been extended to handle also set-based tasks in \cite{Simetti_jirs2016, moe_etal_frontiers2016}.

The choice of the prioritized order of the tasks within the hierarchy has a major importance and strongly affects the behavior of the system, thus it is useful to divide them in three categories and assign them a decreasing priority level \cite{DilArrAntChi_iros18}: safety-related, operational and optimization tasks. Safety tasks such as obstacle avoidance or mechanical joint limits \cite{zanchettin2016safety} have to be necessarily set at an higher priority level with respect to the operational tasks, as they assure the integrity of the system and of the environment in which it operates. At the lowest priority level there are the optimization tasks that help in increasing the efficiency of the operation, but they are are not strictly necessary for its accomplishment.

In this paper we propose a method for increasing the performances of a robotic system by setting proper optimization tasks together with the necessary safety and operational ones. The idea is to set a low-priority optimization task for each one of the safety-related, high-priority task, aiming to minimize the number of transitions between their activation and deactivation states. Given the null-space projection method, the activation of a high-priority task affects the operational task potentially deviating it from the desired value. In this perspective, minimizing the activation of all the safety-related tasks allows the system to better execute the operational task.

Inspired by the work~\cite{sverdrup2017kinematic}, where the set-based multi-task priority framework \cite{moe_etal_frontiers2016} is used to handle, in simulation only, the kinematic singularity of a snake-like robot setting a proper task at two priority level simultaneously, in this work we extend that idea in order to handle several set-based tasks. In addition, we prove its practical efficiency implementing it experimentally on a 7 DOF Jaco$^2$ anthropomorphic arm.

This paper is organized as follows:
    Section \ref{sec:ik} introduces the task priority framework used in the experiments;
    Section \ref{sec:opt} describes the proposed approach for the optimization tasks handling;
     Section \ref{sec:exp} shows the experimental results;
    Section \ref{sec:conc}  presents the conclusions.

\section{Set-Based Task-Priority Inverse Kinematics}\label{sec:ik}
For a general robotic system with $n$ DOF (Degrees of Freedom), the state is described by the joint values $\bfq = \left[q_1,q_2,\dots,q_n\right]^T\in \mathbb{R}^{\it{n}}$. Defining a \textit{task} as a generic $m$-dimensional control objective as a function of the system state $\bfsigma(\bfq) \in \mathbb{R}^{\it{m}}$, the inverse kinematics problem consists in finding the $\bfq$ vector that brings $\bfsigma(\bfq)$ to a desired value $\bfsigma_d$. The  linear mapping between the task-space velocity and the system velocity is \cite{siciliano2010robotics}:
\begin{equation}\label{eq:diffkinematics}
\dot\bfsigma(\bfq)=\bfJ(\bfq)\dot\bfq,
\end{equation}
where $\bfJ(\bfq)= \frac{\partial \bfsigma(\bfq)}{\partial \bfq} \in \mathbb{R}^{\it{m\times n}}$ is the task Jacobian matrix, and $\dot\bfq$ 
is the system velocity vector.  Thus, starting from an initial configuration, in case that $m=n$ meaning that the number of DOF of the system is equal to the task dimension,  the joint increment needed to bring the task value closer to the desired one can be computed by resorting to the CLIK (Closed-Loop Inverse Kinematics) algorithm:
\begin{equation}\label{eq:cliknonredundant}
\dot\bfq = \bfJ^{-1}(\bfq)(\dot\bfsigma_d + \bfK \tilde\bfsigma)
\end{equation}
where $\bfK$ is a positive-definite matrix of gains, $\dot\bfsigma_d$ is the desired task velocity and $\tilde\bfsigma = \bfsigma_d - \bfsigma$ is the task error.

A robotic system is defined redundant if $n > m$, thus if it has more DOF that the required ones for the accomplishment of a certain task. In this case the Jacobian matrix is not invertible anymore and multiple possible solutions for Eq. \ref{eq:diffkinematics} exist. The system can be solved imposing a constrained minimization problem, in which the cost function is:
\begin{equation}\label{eq:constraint}
g(\dot\bfq) = \frac{1}{2} \dot\bfq^T \dot\bfq
\end{equation} 
that selects among all the possible solutions the one that minimizes the joint velocity norm. In this way Eq. \ref{eq:cliknonredundant} becomes:
\begin{equation}\label{eq:clikredundant}
\dot\bfq = \bfJ^\dagger(\bfq)(\dot\bfsigma_d + \bfK \tilde\bfsigma)
\end{equation}
where $\bfJ^\dagger$ is the Moore-Penrose pseudoinverse of the Jacobian matrix, defined as:
\begin{equation}
    \bfJ^\dagger = \bfJ^T (\bfJ \bfJ^T)^{-1}
\end{equation}

In general the solution of Eq. \ref{eq:clikredundant} lies is the subspace $\mathcal{R}(\bfJ)$, and in the redundant case its orthogonal complement $\mathcal{N}(\bfJ) \neq \emptyset$ can be exploited to add other components that would not affect the accomplishment of the task. For this reason the general solution can be written as:
\begin{equation}\label{eq:generalsolution}
\dot\bfq = \bfJ^\dagger \dot\bfsigma + \bfN \dot\bfq_0
\end{equation}
where:
$$
\bfN = \bfI - \bfJ^\dagger \bfJ
$$
is the null space projector and $\dot\bfq_0$ is an arbitrary vector that can be used to minimize or maximize a scalar value by setting
$$
\dot\bfq_0 = k_0 (\frac{\partial w(\bfq)}{\partial \bfq})^T
$$
where $k_0$ is a gain and $w(\bfq)$ is a secondary objective function.

Another useful exploitation is to define a second task with a specific desired value and compute the solution that accomplishes both the tasks. Unfortunately this solution may not exist due to the infeasibility of their simultaneous resolution. In this case it is necessary to define a priority between the tasks and compute the solution that achieves the primary one while minimizes the error on the secondary one. Given two tasks $\bfsigma_1$ and $\bfsigma_2$ it is possible to compute the system velocities $\dot\bfq_1$ and $\dot\bfq_2$ that accomplish them separately using Eq. \ref{eq:clikredundant}.
The composition of the two tasks 
solutions $\dot\bfq_1$ and $\dot\bfq_2$ can be performed by resorting to the SRMTP (Singularity-Robust Multi-Task Priority) framework \cite{Chi_tra97}:
\begin{equation}\label{eq:ikchiaverini}
\dot\bfq = \dot\bfq_1 + \bfN_1 \dot\bfq_2
\end{equation}

In \cite{AntArrChi_paladyn2010,AntArrChi_ISR2008} the SRMTP Inverse Kinematics framework has been extended to handle an arbitrary number of tasks, 
by resorting to the NSB (Null Space-based Behavioral) control. Given a hierarchy composed by $h$ tasks sorted by priority level, the solution is computed as:
\begin{equation}\label{eq:nsb}
\dot\bfq = \dot\bfq_1 + \bfN_1^A \dot\bfq2 + \cdots + \bfN_{h-1}^A \dot\bfq_h
\end{equation}
where $N_i^A$ is the null space projector of the augmented Jacobian matrix $J_i^A$ defined
as:
\begin{equation}\label{eq:augmented}
\bfJ_i^A = \begin{bmatrix}
\bfJ_1^T &
\bfJ_2^T &
\hdots &
\bfJ_i^T
\end{bmatrix}^T
\end{equation}

The NSB algorithm has been developed to handle {\textit{equality-based}} tasks, thus control objectives in which the goal is to bring the task value to a specific one, e.g. moving the arm end-effector to a target position and orientation. However, several control objectives may require their value to lie in an interval, i.e. above a lower threshold and below an upper threshold. These are usually called \textit{set-based} tasks or inequality constraints.
A set-based task can be seen as an equality-based one which gets active or inactive depending on the operational conditions. In particular, it is necessary to set different reference values for each set-based task: physical thresholds $\sigma_{M}$ ($\sigma_{m}$), safety thresholds $\sigma_{s,u}~<~\sigma_M$ ($\sigma_{s,l}~>~\sigma_m$), and activation thresholds $\sigma_{a,u} =  \sigma_{s,u}~-~\epsilon$ ($\sigma_{a,l}~=~\sigma_{s,l}~+~\epsilon$). When the task value reaches an activation threshold, it is added to the task hierarchy as a new equality-based task with desired value equal to the corresponding safety threshold:
 \begin{equation}
\sigma_d=\left\{\begin{array}{l l}
\sigma_{s,u} &\text{if $\sigma\geq\sigma_{a,u}$}\\
\sigma_{s,l} &\text{if $\sigma\leq\sigma_{a,l}$}
 \end{array} \right.
\end{equation}

Then it can be deactivated when the solution of the hierarchy that contains only the other tasks would push its value toward the valid set. Defining $\bfJ_A$ as the Jacobian matrix of $\sigma_A$, if $\bfJ_A \dot\bfq > 0$ the solution would increase the set-based task value, otherwise if $\bfJ_A \dot\bfq < 0$ the solution would decrease it. In this way, $\sigma_A$ can be deactivated if

\begin{equation}\label{eq:cond1}
\sigma_A \geq \sigma_{a,u} \; \land \; \bfJ_A \dot\bfq < 0
\end{equation}
or
\begin{equation}\label{eq:cond2}
\sigma_A \leq \sigma_{a,l} \; \land \; \bfJ_A \dot\bfq > 0
\end{equation}

\begin{psfrags}
\psfrag{Active}[1pt][1pt][0.6]{Active}
\psfrag{Inactive}[1pt][1pt][0.6]{Inactive}

\psfrag{sigmal}[1pt][1pt][0.8]{$\sigma_{a,l}$}
\psfrag{sigmasl}[1pt][1pt][0.8]{$\sigma_{s,l}$}
\psfrag{sigmau}[1pt][1pt][0.8]{$\sigma_{a,u}$}
\psfrag{sigmasu}[1pt][1pt][0.8]{$\sigma_{s,u}$}
\psfrag{task value}[1pt][1pt][0.8]{task value}

\mypsfrag{8}{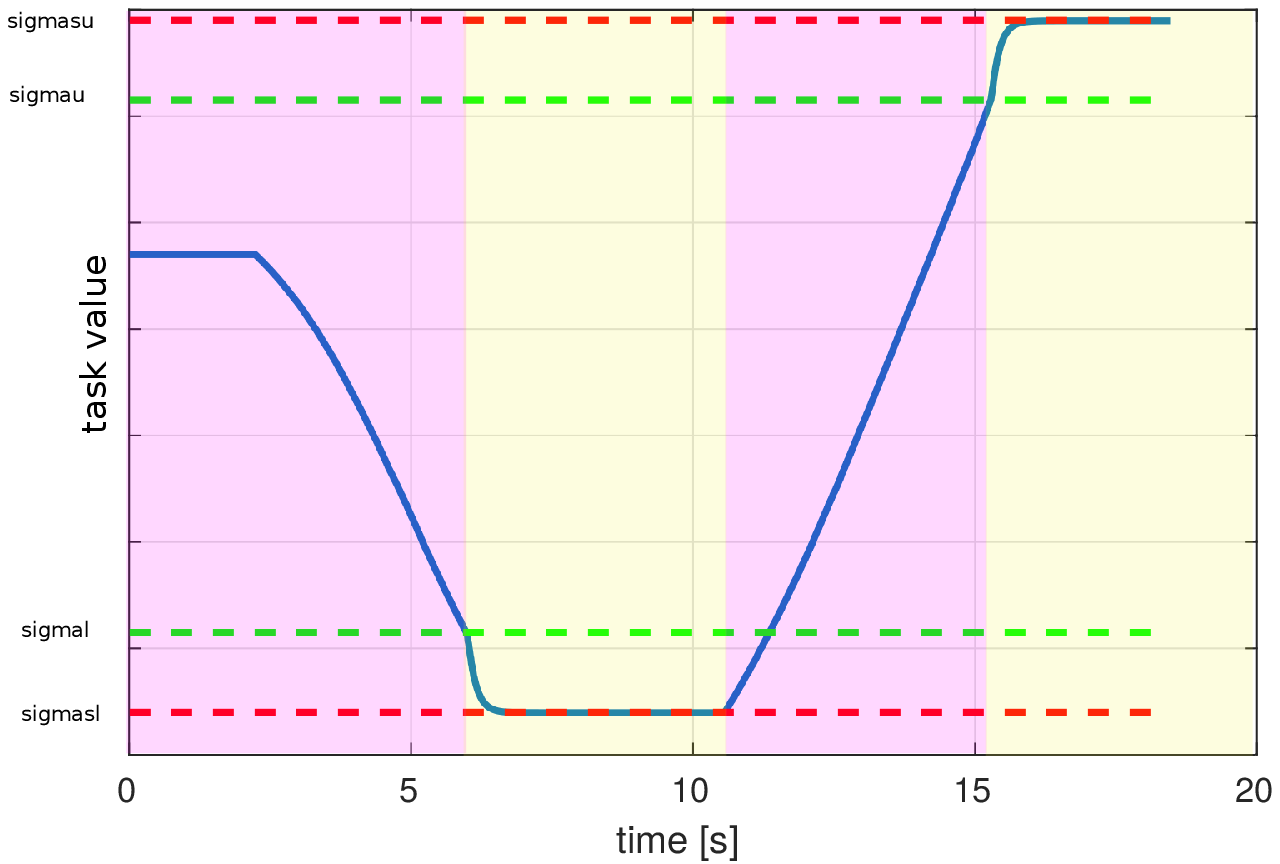}{-12pt}{A generic set-based task value over time, the corresponding $\sigma_{a,l}$, $\sigma_{a,u}$ (green dashed lines) and $\sigma_{s,l}$, $\sigma_{s,u}$ (red-dashed lines), activation (magenta background) and deactivation (yellow background) state.}{fig:example}
\end{psfrags}

Figure \ref{fig:example} shows a generic set-based task value over time and the corresponding $\sigma_{a,l}$, $\sigma_{a,u}$ (green dashed lines) and $\sigma_{s,l}$, $\sigma_{s,u}$ (red-dashed lines). The background color highlights the activation (magenta) and deactivation (yellow) state. 
At the beginning, the task is inactive, as its value lies within the valid interval, thus the hierarchy contains 
only the other tasks.
As soon as its value reaches $\sigma_{a,l}$, it gets activated and added to the current hierarchy. The corresponding solution brings the task value to $\sigma_{s,l}$ with a time constant proportional to the task gain. It remains at that threshold until condition (\ref{eq:cond1}) or (\ref{eq:cond2}) are satisfied. From that point the task is deactivated and it is removed from the hierarchy. The same happens with the upper thresholds $\sigma_{a,u}$ and $\sigma_{s,u}$. For more details about the activation/deactivation algorithm see \cite{arrichiello2017assistive}.

\section{Optimization tasks handling}\label{sec:opt}
The priority order among the tasks in the hierarchy strongly affects the behavior of the system during the execution of a certain operation. All the set-based tasks related to the safety of the system, such as the mechanical joint limits and the obstacle avoidance, have to be necessarily placed at the top priority level. The execution of the operational task is constrained to the fulfillment of an active higher-priority task and, in case of conflict between them, it leads to a deviation of the operational task from the desired trajectory. Optimization tasks such as the maximization of the arm manipulability can be placed at a lower priority level with respect to the operational one, due to the fact that they are not strictly necessary for the accomplishment of the operation. The idea that we propose in this work is to define proper optimization tasks aiming to minimize the activation/deactivation of high-priority safety tasks. In particular, it would be desirable that the control algorithm tries to push a high-priority task further away from the imposed minimum/maximum limits even when it is not active, exploiting the system redundancy. In order to implement this method in the Set-Based Multi-Task Priority framework an equality-based optimization task should be added in the hierarchy for each one of the set-based high-priority tasks, at a low-priority level with desired value:
\begin{itemize}
    \item greater than the maximum task value if the corresponding set-based task has a lower threshold
    \item lower that the minimum task value if the corresponding set based task has an upper threshold
    \item equal to the mid-point between the minimum and maximum thresholds if the corresponding set-based task has both of them
\end{itemize}
obtaining the hierarchy shown in Fig. \ref{fig:priorita}. This kind of approach leads to the minimization of the high-priority tasks activation and improves the system performances in tracking the operational task, always assuring that the safety thresholds are respected during the motion. 
\begin{psfrags}
\psfrag{s1}[1pt][1pt][1.0]{safety task $\bfsigma_1$}
\psfrag{s2}[1pt][1pt][1.0]{safety task $\bfsigma_2$}
\psfrag{sn}[1pt][1pt][1.0]{safety task $\bfsigma_n$}
\psfrag{p1}[1pt][1pt][1.0]{operational task $\bfsigma_A$}
\psfrag{so1}[1pt][1pt][1.0]{optimization task $\bfsigma_1^o$}
\psfrag{so2}[1pt][1pt][1.0]{optimization task $\bfsigma_2^o$}
\psfrag{son}[1pt][1pt][1.0]{optimization task $\bfsigma_n^o$}
\mypsfrag{5}{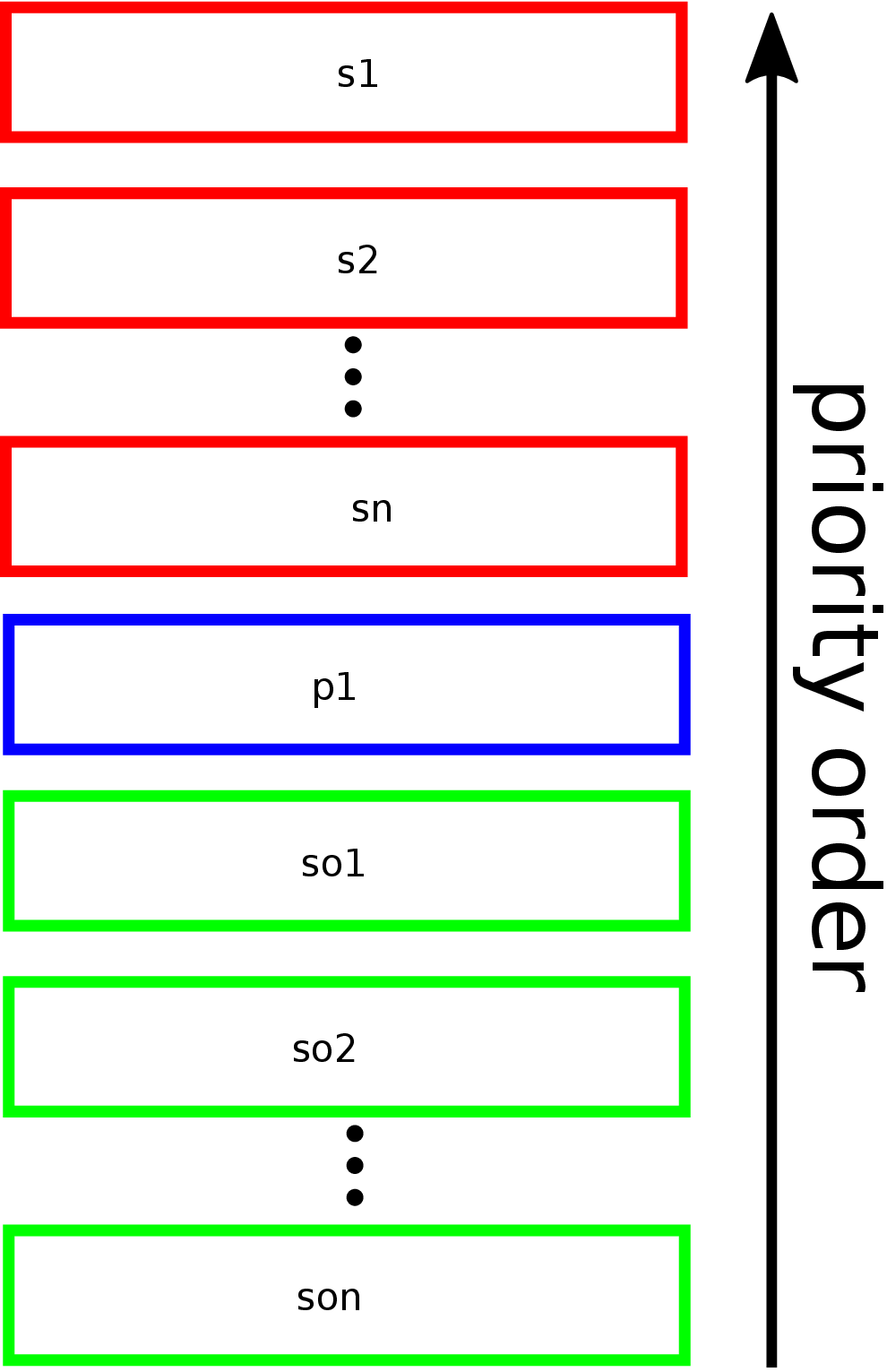}{-4pt}{Proposed task hierarchy: for each one of the high-priority safety tasks there is the corresponding low-priority optimization task aimed to minimize its activations}{fig:priorita}
\end{psfrags}

In this work we take into account two kind of tasks: the arm manipulability and the joint limits.
The measure of manipulability \cite{yoshikawa1985manipulability}:
$$
\sigma(\bfq) = \sqrt{ \text{det}(\bfJ \bfJ^T)}
$$
 goes to zero when the manipulator reaches a singular configuration, as $\bfJ$ loses rank. For this reason it can be seen as a {\textit{distance}} from a singular configuration.
 It is possible to add it to a hierarchy as a high-priority set-based task, defining a minimum threshold that the manipulator cannot exceed during the movement. For this work the task Jacobian is computed numerically following the Algorithm \ref{alg:manip}. 
\begin{algorithm}\label{alg:manip}
\SetAlgoLined
\KwData{current joint positions vector $\bfq \in \mathbb{R}^n$ }
\KwResult{numeric Jacobian of the manipulability task $\bfJ \in \mathbb{R}^{1\times n}$ }
initialize $\Delta q$, $\bfq_\text{inc}$\\
\For{i=1:$n$}{
\For{j=1:$n$}{
\eIf{ j=i}{
  $q_\text{inc}(j) = q(j) + \Delta q$\\
}
{$q_\text{inc}(i)=q(j)$}
}
$w=\;$ManipulabilityValue($\bfq$) \\
$w_\text{inc}=\;$ManipulabilityValue($\bfq_\text{inc}$) \\
$J(1,i) = (w_\text{inc}-w)/\Delta q$
}
\caption{Computing the manipulability Jacobian}
\end{algorithm}
 Setting it at a lower priority with respect to the operational task implies that it is accomplished only when it is not in conflict with the primary task, and
it can be used as a maximization control objective: choosing the desired value higher than the maximum measure of manipulability that the arm can exhibit 
  the resulting behavior is the same as applying Eq. \ref{eq:generalsolution}, thus the arm follows the desired trajectory trying to maximize the manipulability measure.
 
 Merging the aforementioned behaviors by including in the hierarchy two manipulability tasks, one at lower priority and one at a higher priority with respect to the operational one,
 the resulting behavior 
is that the arm never reaches a singular configuration (for the effect of the high-priority task) while tries to maximize the manipulability measure
during all the trajectory even when the corresponding set-based task in inactive (for the effect of the low-priority task), without interfering with the operational one. 

The joint limit task is usually used for avoiding self-collisions, and it can be seen as a high-priority set-based task with a lower threshold  $\sigma_{s,l}$ and an upper threshold $\sigma_{s,u}$ that constraints its movement in a feasible set of values. The task value is simply the $i$-th joint position while the Jacobian is a row vector with a $1$ at the $i$-th column and zeros at the other ones. The corresponding optimization task is an equality-based task in which the desired value is:
$$
\sigma_d = \frac{\sigma_{s,u} + \sigma_{s,l}}{2}
$$
In this way the joints are pushed toward the middle of the chosen limits while the end-effector follows the desired trajectory, minimizing the activation of the high-priority set-based task.
\section{Experimental results}\label{sec:exp}
In this section experimental results on a 7DOF Kinova Jaco$^2$ manipulator that prove the effectiveness of the proposed approach are shown. In the first case study we focus the attention on the joint limit tasks, while the second one takes into account the manipulability tasks. In both case studies we first perform the experiment with the hierarchy that contains only the high-priority set-based tasks, followed by the corresponding experiment in which we add also the low-priority optimization tasks.

\subsection{First case study: joint limits task}
\begin{psfrags}
\mypsfrag{8}{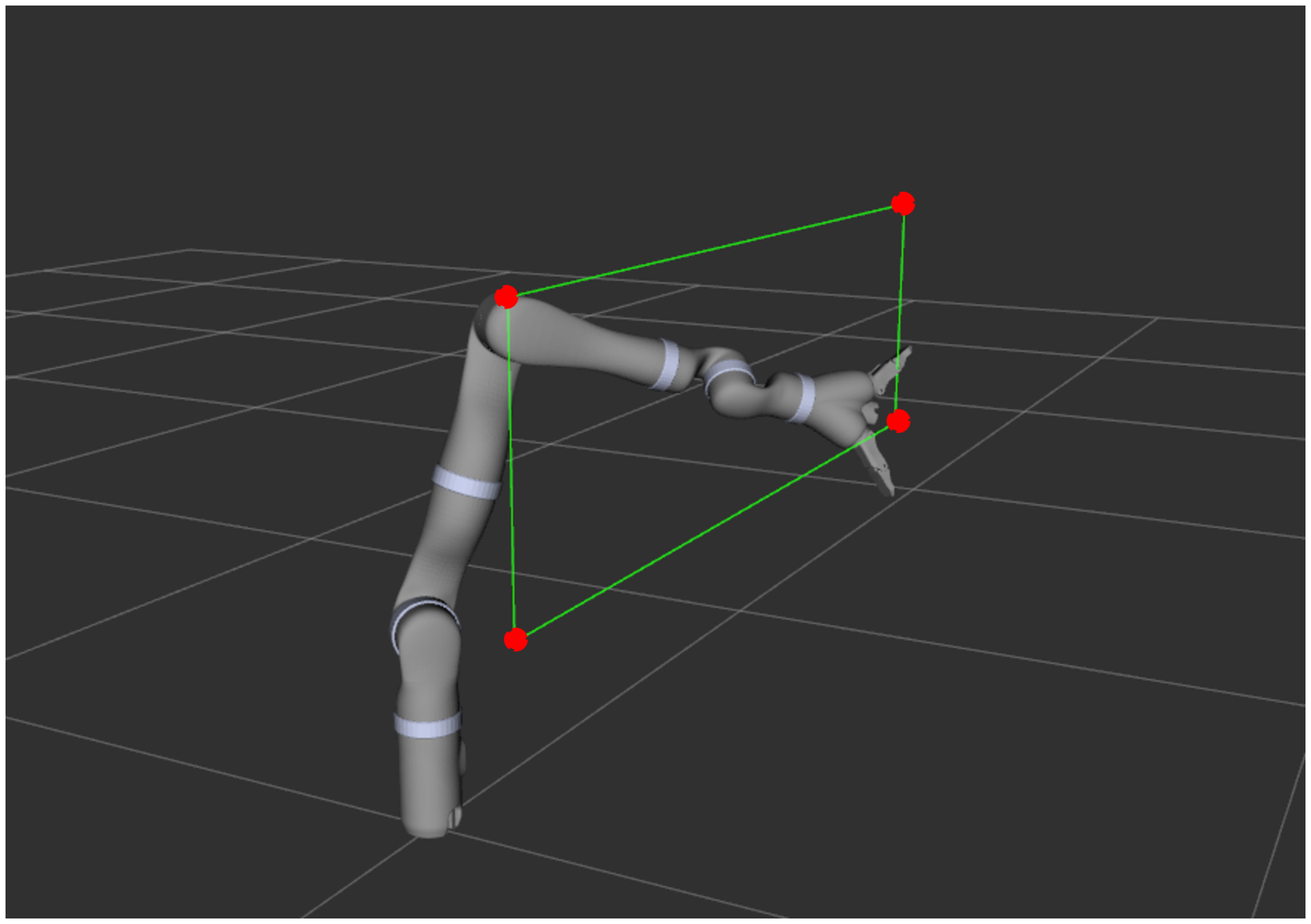}{-4pt}{Desired path for the first case study represented in Rviz}{fig:manip-trajectory}
\end{psfrags}
The desired path for the end-effector is shown in Fig. \ref{fig:manip-trajectory}.
It is composed by four waypoints with a square shape keeping constant the $x$ coordinate of the arm base frame. Upper and lower limits, intentionally chosen in order to get active during the motion, on six joints have been set and added at a high priority with respect to the end-effector position task, obtaining the following hierarchy:
\begin{enumerate}
    \item Joint limits (set based)
    \item End-effector position (equality)
\end{enumerate}
Figure \ref{fig:provarow1} shows the joint positions over time together with the imposed limits. Notice that the seventh joint position is not reported because it is not included in the hierarchy as it does not give any contribution to the position of the end-effector.
\begin{psfrags}
\mydoublerowpsfrag{8}{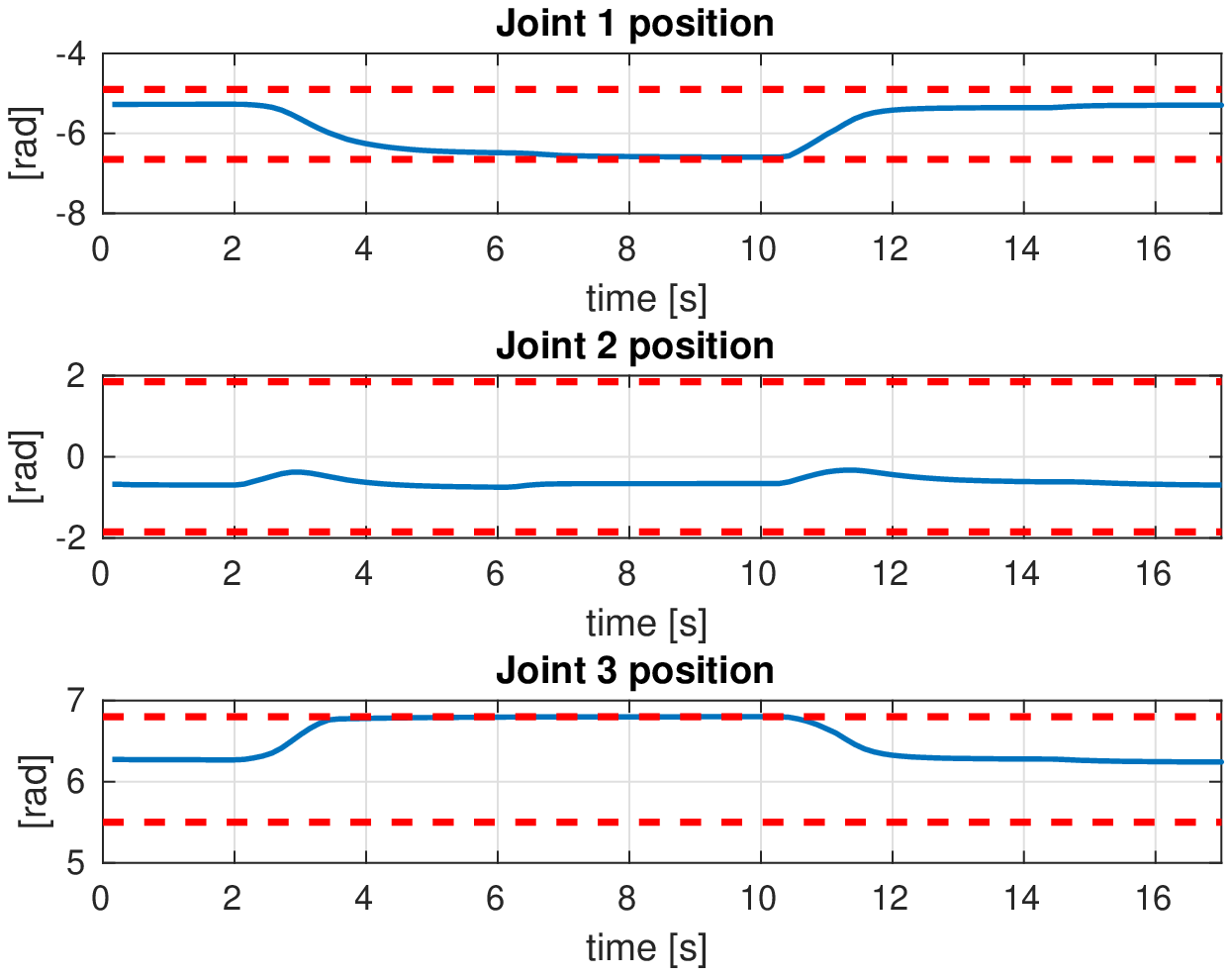}{8}{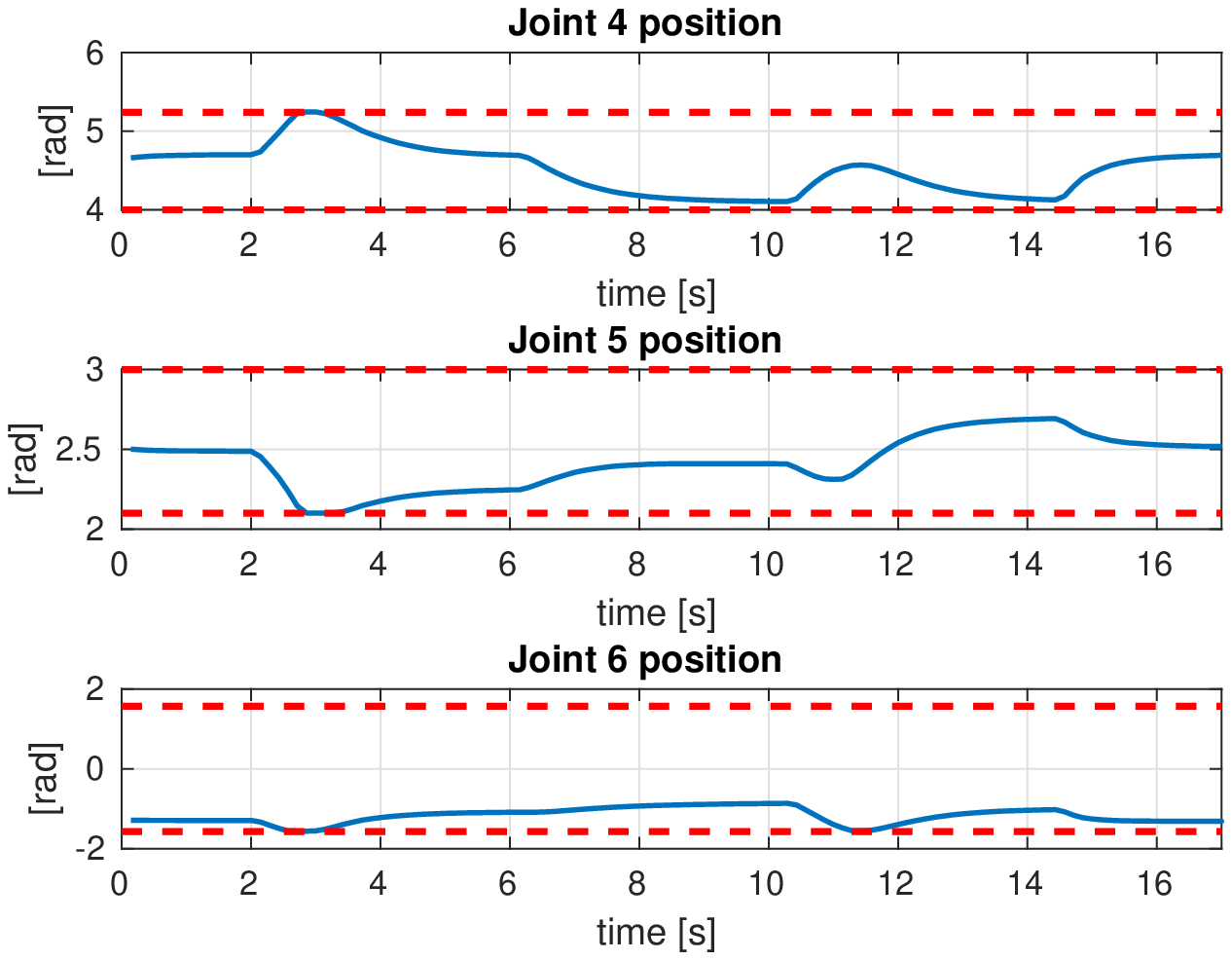}{First case study, only high-priority joint limit tasks: joint positions over time and safety thresholds (red-dashed lines). Five joints reach the limits during the execution of the trajectory.}{fig:provarow1}
\end{psfrags}
It is clear that the joints 1,3,4,5 and 6 reach one of the limits during the movement, activating the corresponding task that stops their motion.

For the second experiment the implemented hierarchy is:
\begin{enumerate}
    \item High-priority joint limits (set based)
    \item End-effector position (equality)
    \item Low-priority joint optimization (optimization)
\end{enumerate}
The starting configuration, the desired sequence of waypoints and the the imposed joint limits are the same of the previous experiment and Fig. \ref{fig:provarow2}  shows the joint positions.
\begin{psfrags}
\mydoublerowpsfrag{8}{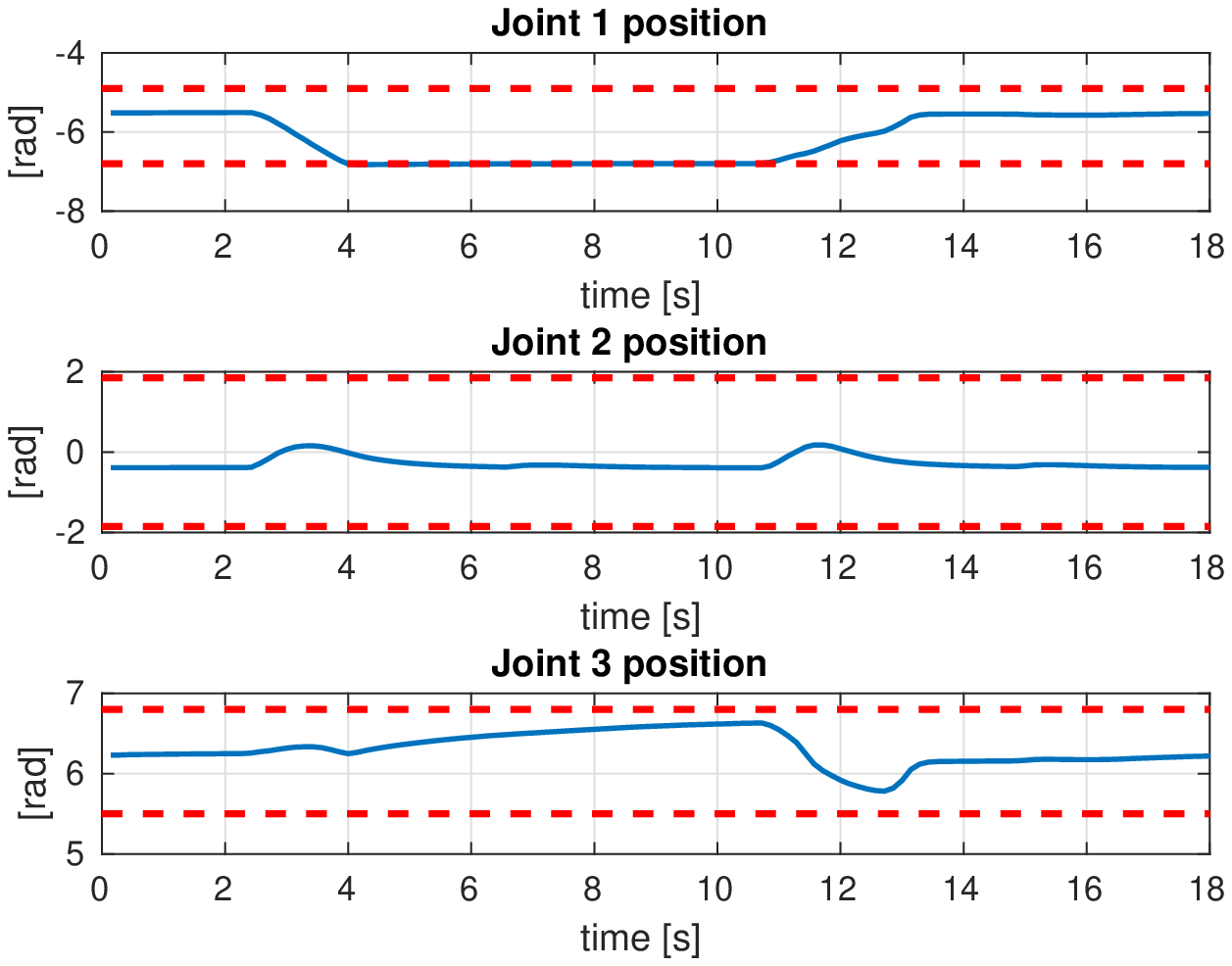}{8}{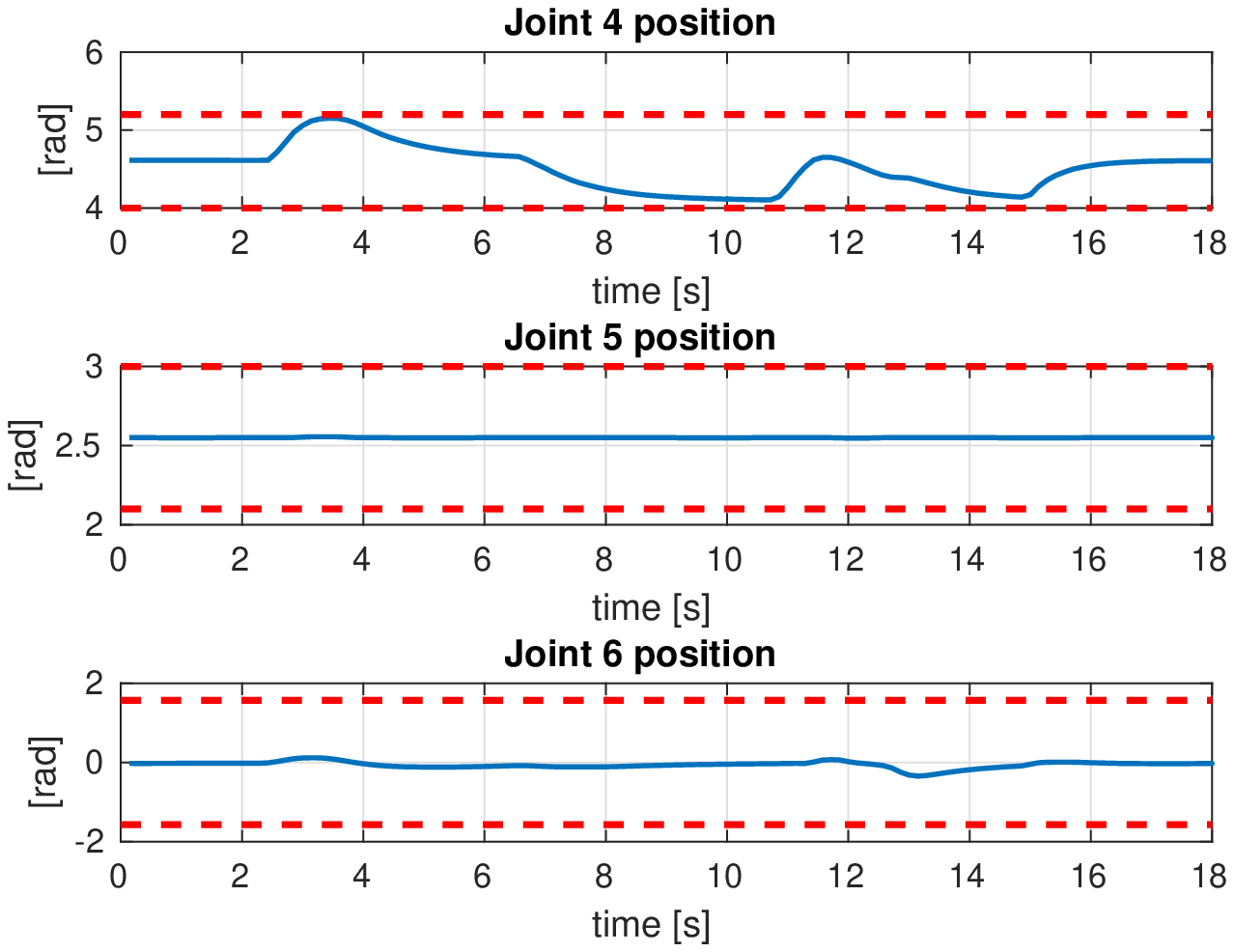}{First case study, high-priority and optimization joint limit tasks: joint positions over time and safety thresholds (red-dashed lines). The optimization tasks make the joints stay further from the limits with respect to the previous experiment.}{fig:provarow2}
\end{psfrags}
It is possible to notice that this time only two joints reach the limits, while joints 3, 5 and 6 benefit from the lower-priority optimization tasks, being further from the limits with respect to the previous experiment. Figure \ref{fig:prova01-3} shows a 3D representation of the executed path for the two experiments. The activation of the joint limit tasks in the first experiment makes the end-effector deviate from the desired path, while in the second one it follows the desired path clearly better given that less higher priority tasks get active during the movement. The end-effector does not track perfectly the desired path because two joint limits get active anyway.  
\begin{psfrags}
\mypsfrag{8}{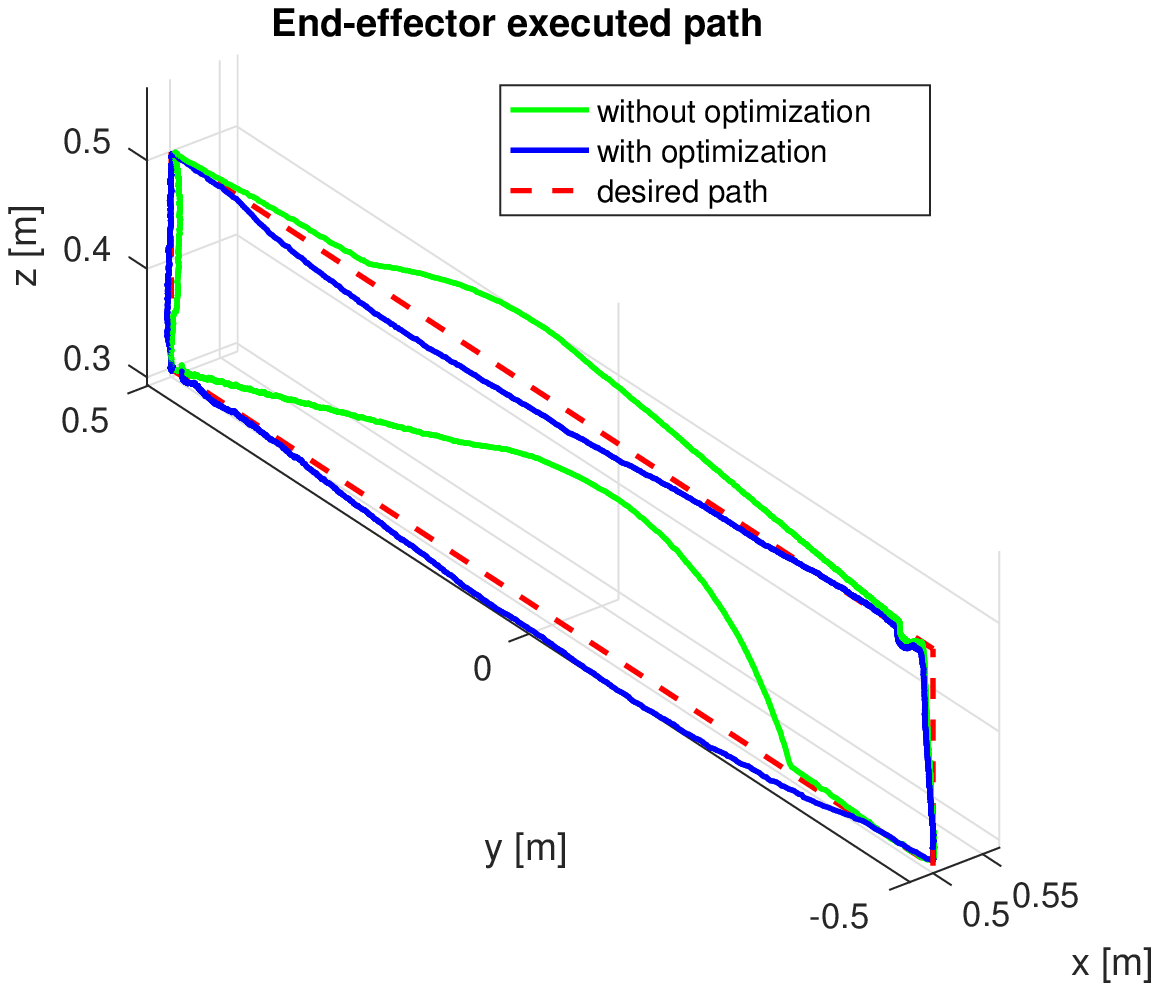}{-4pt}{First case study: comparison between the executed path obtained with and without the optimization tasks (blue and green lines respectively) and desired path (red-dashed line). The executed path  obtained adding the optimization tasks tracks better the reference, given the less frequent activation of the safety tasks.}{fig:prova01-3}
\end{psfrags}
\subsection{Second case study: manipulability task}
The second case study takes into account the behavior of the system when two manipulability tasks are added to the hierarchy with different priority order. The desired path for the end-effector is shown in Fig. \ref{fig:disegno}. 
\begin{psfrags}
\mypsfrag{7}{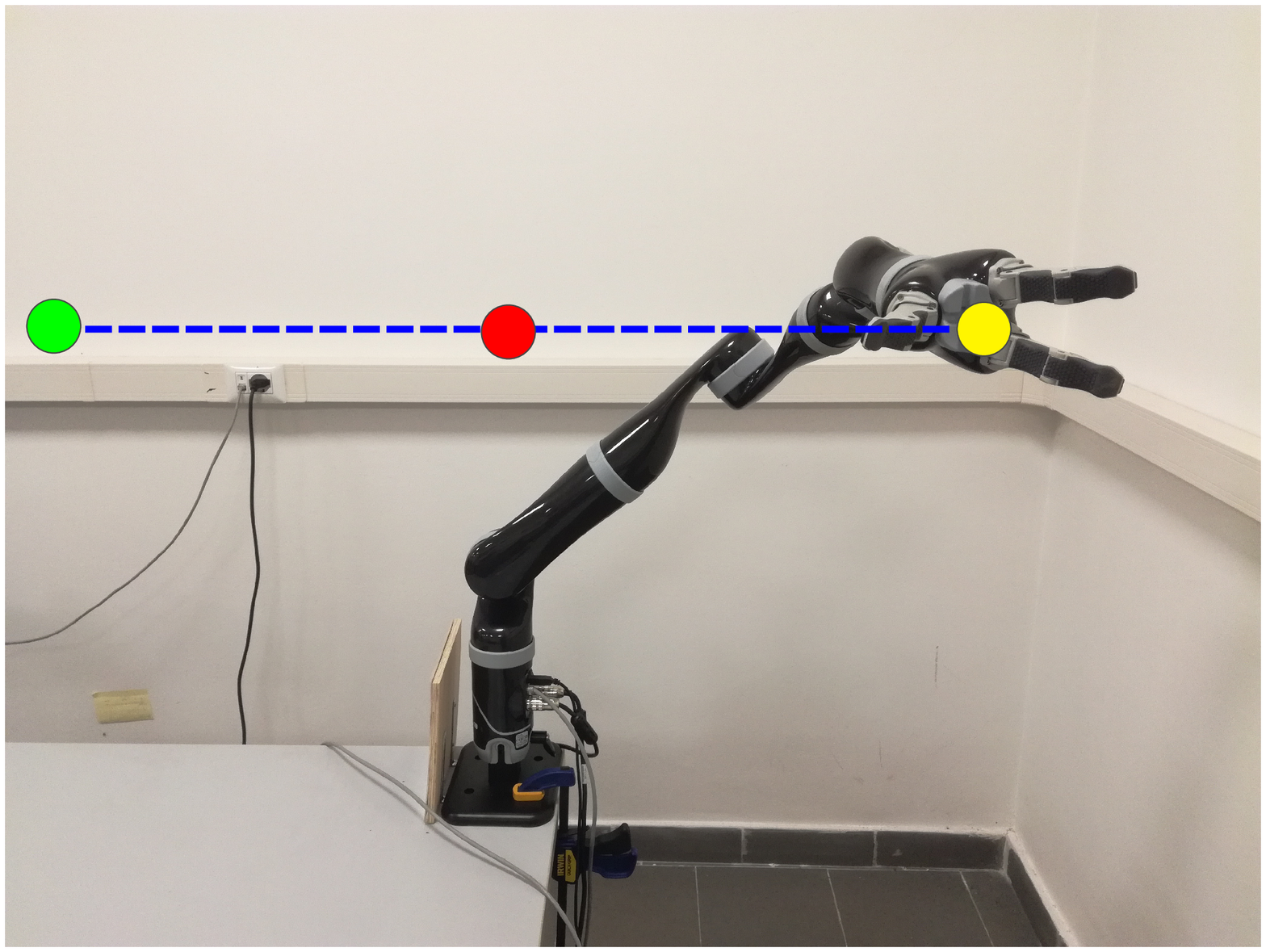}{-4pt}{Desired path for the second case study. It is composed by two waypoints (yellow and green circles). The red circle is associated with a configuration in which the measure of manipulability is very low ($10e^{-5}$)}{fig:disegno}
\end{psfrags}
It is a simple straight line (blue-dashed) that starts from the yellow circle and ends on the green circle, to be followed forwards and backwards with a constant orientation. The red circle corresponds to the configuration depicted in Fig. \ref{fig:singolare}, in which the measure of manipulability reaches a very low value ($10^{-5}$).
\begin{psfrags}
\mypsfrag{7}{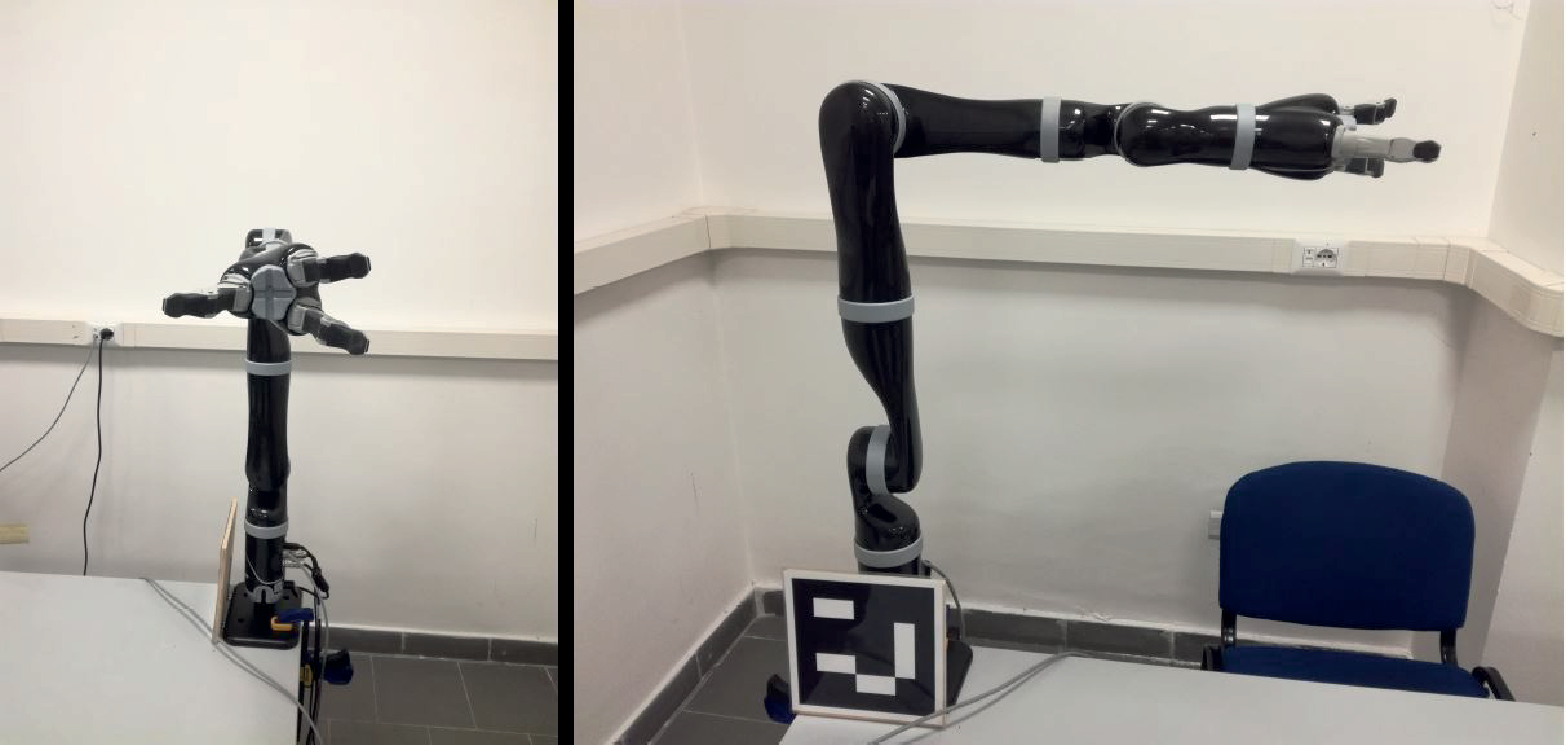}{-4pt}{Configuration close to singularity. Front view (left) and lateral view (right)}{fig:singolare}
\end{psfrags}
In the first experiment the task hierarchy is:
\begin{enumerate}
\item High-priority arm manipulability (set based)
\item End-effector position and orientation (equality)
\end{enumerate}
Figure \ref{fig:exp03-1} shows the measure of manipulability over time. The task gets active two times, when the desired trajectory reaches the red circle, and the control algorithm effectively avoids the singular configuration keeping the measure of manipulability above the  chosen threshold.
\begin{psfrags}
\mypsfrag{8}{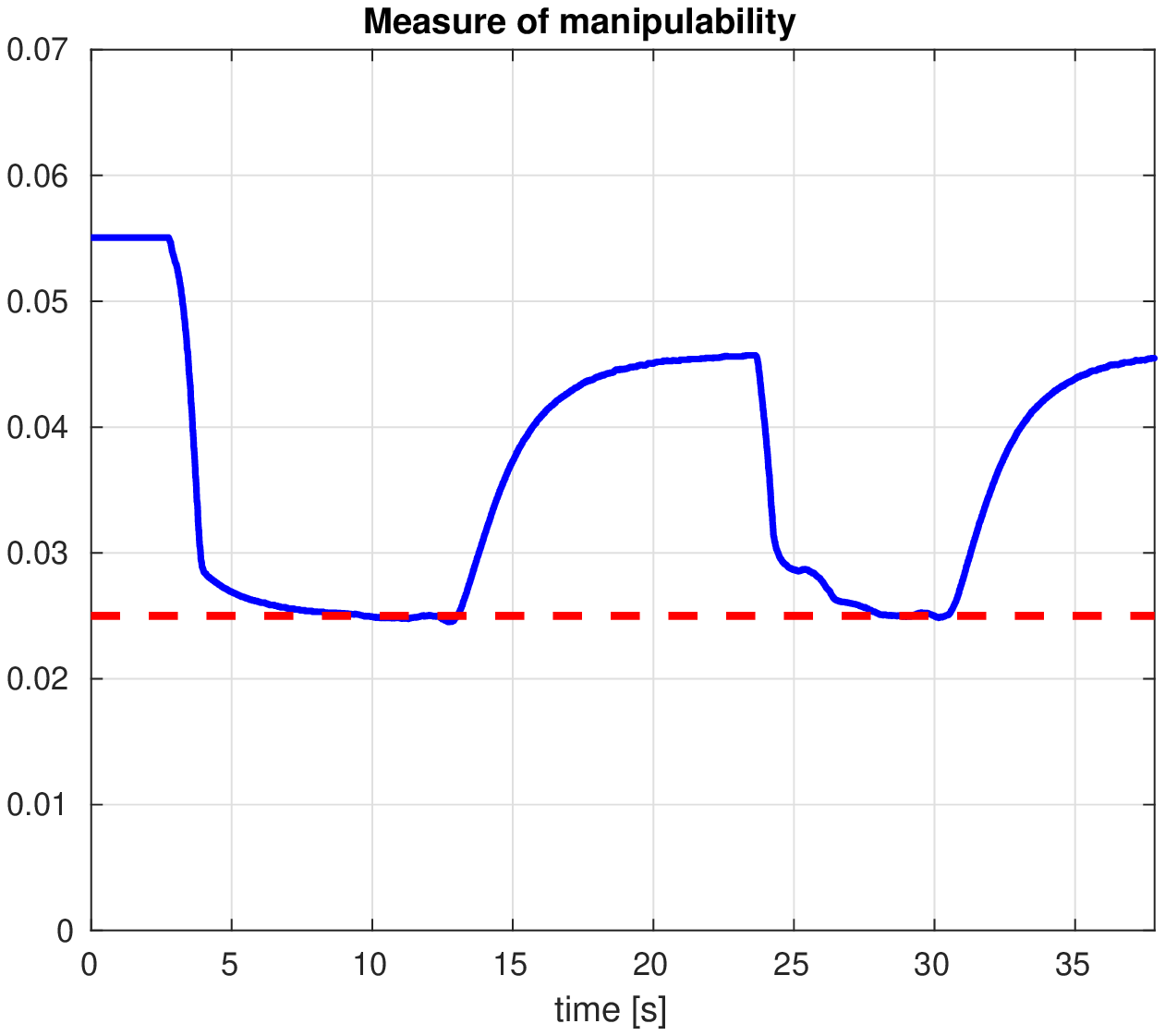}{-4pt}{Second case study, only high-priority manipulability task. Measure of manipulability over time and the imposed safety threshold (red-dashed). The task gets active two times and its value never exceeds the chosen safety threshold.}{fig:exp03-1}
\end{psfrags}

Let us now add a second manipulability task at a lower priority while following the same path, resulting in the following hierarchy:
\begin{enumerate}
\item High-priority arm manipulability (set based)
\item End-effector position and orientation (equality)
\item Low-priority arm manipulability (optimization) 
\end{enumerate}
Figure \ref{fig:exp04-2} shows the measure of manipulability with the chosen threshold for the primary manipulability task. The desired value for the low-priority manipulability task is set at $1.2$, which is greater than the maximum value that the arm can exhibit.
\begin{psfrags}
\mypsfrag{8}{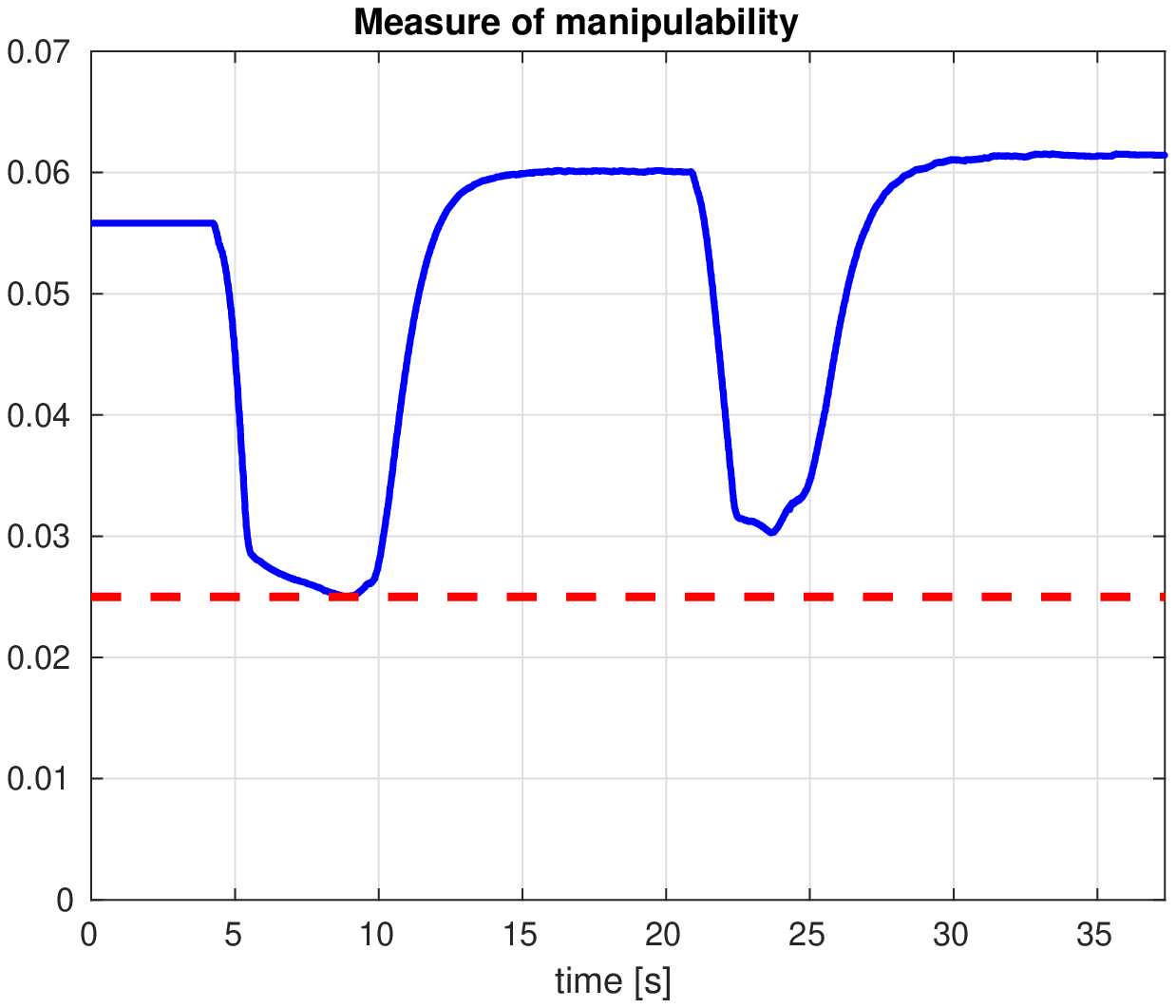}{-4pt}{Second case study, high-priority and optimization manipulability tasks: measure of manipulability over time and imposed safety threshold (red-dashed) of the primary manipulability task. The high-priority one gets active only once during the experiment, given the maximization of the measure of manipulability during all the movement, even when the it is not active. }{fig:exp04-2}
\end{psfrags}
It is clear that the lower priority manipulability task maximizes the value: the result is that when the desired trajectory reaches for the first time the singular configuration the corresponding task gets active, but it deactivates very quickly with respect to the previous case. Additionally, when the end-effector returns to the initial position the task does not even activate, because the maximization of the manipulability measure during the trajectory has rearranged the configuration of the arm, basically changing the position of the elbow. Figure \ref{fig:exp04-1} shows a comparison of  the second part of the executed path between the two experiments, from the green circle to the yellow circle.
\begin{psfrags}
\mypsfrag{8}{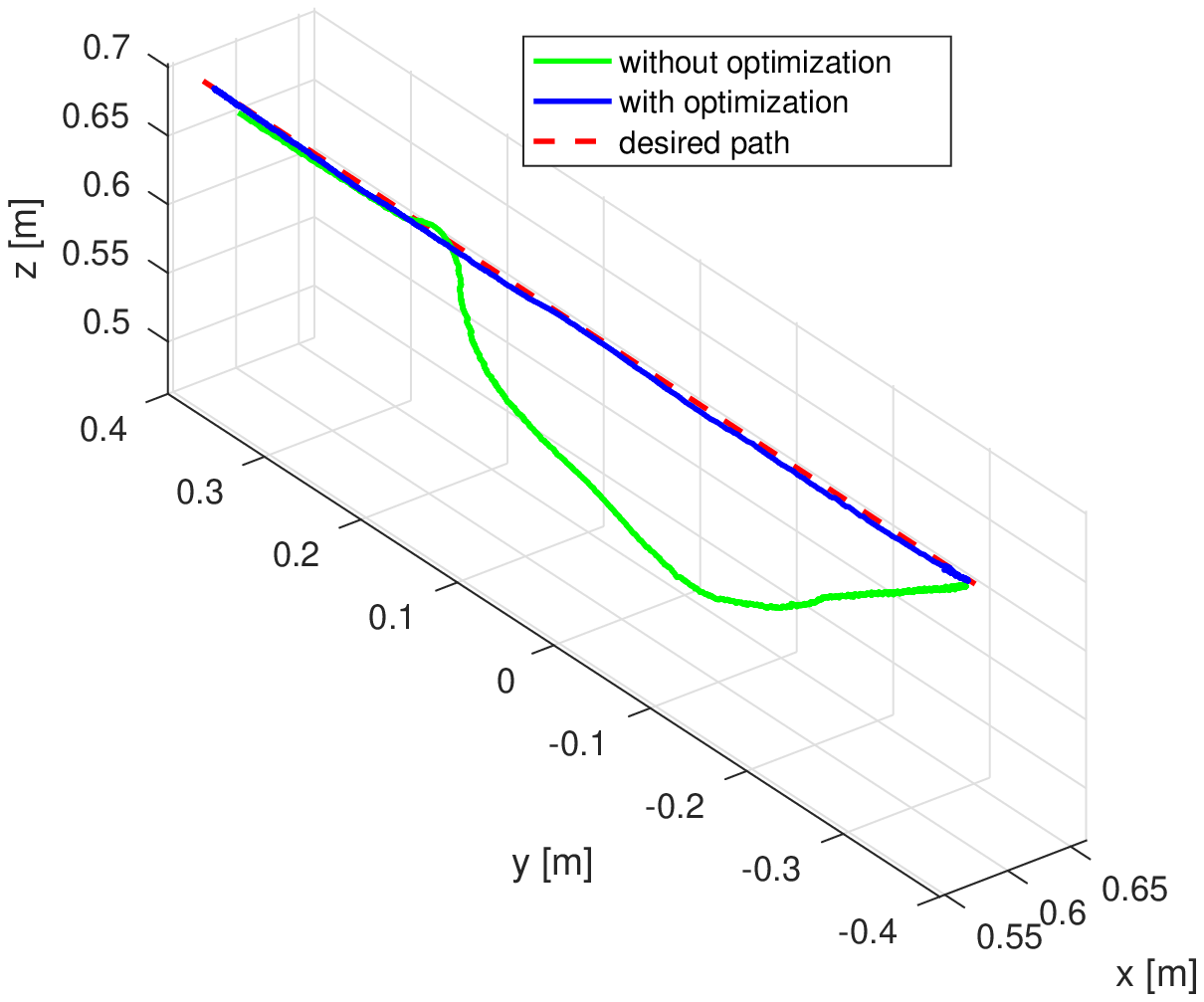}{-4pt}{Second case study: second part of the executed path obtained with and without the optimization tasks (blue and green lines respectively) and desired path (red-dashed line). The executed path  obtained adding the optimization tasks tracks better the reference, because the high-priority task does not get active during this part of the movement.}{fig:exp04-1}
\end{psfrags}
In the experiment performed with the hierarchy containing only the high-priority manipulability task the executed path deviates from the desired one given the activation of the set-based task. In the second case it does not get active, and the end-effector can track the desired path much better.
\section{Conclusions}\label{sec:conc}
In this paper we have shown a method for improving the tracking capabilities of the operational tasks in presence of higher-priority safety tasks in the hierarchy. We have first described the inverse kinematics framework that allows to define different kind of tasks and to sort them in priority. Then we have discussed how the choice of proper optimization tasks at a lower priority level for each one of the safety tasks can minimize the activation of the safety-tasks, leading to a better execution of the operational ones. Experimental results on a 7DOF Kinova Jaco$^2$ arm proved the effectiveness of the proposed method on two different kind of set-based safety tasks.  

\section*{Acknowledgments}
This work was supported by the European Community through the project AEROARMS(H2020-635491).

\bibliography{example}

\end{document}